\documentclass[runningheads]{llncs}
\usepackage[T1]{fontenc}
\usepackage{graphicx}
\usepackage{amsmath}
\usepackage{comment}
\usepackage[table]{xcolor}
\usepackage{adjustbox}
\usepackage{xcolor}
\usepackage{booktabs}
\usepackage{amssymb}
\usepackage[normalem]{ulem}
\usepackage[linesnumbered,ruled,vlined]{algorithm2e}
\SetAlgoLined
\DontPrintSemicolon
\usepackage{upgreek}
\usepackage{bm}
\usepackage[table]{xcolor}
\usepackage[linesnumbered,ruled,vlined]{algorithm2e}
\usepackage[table]{xcolor}
\usepackage[linesnumbered,ruled,vlined]{algorithm2e}

\begin{document}
\title{Adaptive Weighted LS-SVM for Multi-View Classification}
%
%\titlerunning{Abbreviated paper title}
% If the paper title is too long for the running head, you can set
% an abbreviated paper title here
%
% \author {Anonymous submission}
\author{Farnaz Faramarzi Lighvan \and
Mehrdad Asadi\and
Lynn Houthuys}
\authorrunning{F.F Lighvan et al.}

% First names are abbreviated in the running head.
% If there are more than two authors, 'et al.' is used.
%

\institute{AI Lab, Vrije Universiteit Brussel, Belgium}
\maketitle              % typeset the header of the contribution

\begin{abstract}
Multi-view learning integrates diverse representations of the same instances and can improve performance when interactions across views are effectively exploited. Most existing kernel-based multi-view learning methods either rely on fusion techniques without explicitly enforcing a consensus or complementary collaboration across views, or use co-regularization-based loss functions that impose only pairwise interactions, thereby limiting global collaboration. We propose AW-LSSVM, an adaptive weighted LS-SVM that explicitly enforces complementary learning across all views through an iterative global coupling mechanism. At each iteration, each view not only learns from its own data but is also guided to compensate for samples misclassified by other views in previous iterations by assigning adaptive sample weights. We introduce two strategies for computing these weights: (1) based on averaging misclassification errors across other views and, (2) based on a dissimilarity-aware error aggregation that puts more emphasis on errors from more dissimilar views. Experiments demonstrate that AW-LSSVM outperforms existing multi-view methods on most benchmark datasets. %Moreover, its ability to keep the raw features of each view isolated makes it compatible with privacy-sensitive or distributed multi-view settings.

\end{abstract}
\section{Introduction}
Multi-view learning integrates multiple representations of the same data to improve performance. Views may come from different data types or feature groups and provide correlative or complementary information. Correlative views enhance generalization, while complementary ones lead to more robust models. A core component of multi-view learning is the fusion strategy~\cite{fusion}. Early fusion merges input-level features for joint modeling, but can be vulnerable to noise and prone to issues arising from high dimensionality. Late fusion combines predictions from separate models but may miss inter-view interactions. Mid fusion aims to balance both approaches.

Kernel-based methods are commonly used in multi-view classification tasks due to their capacity for nonlinear modeling, strong generalization performance on small datasets, and support for flexible regularization. Multiple kernel learning~\cite{aiolli2015easymkl} aggregates kernels from each view to learn a shared decision function without explicitly enforcing a specific type of collaboration among views. Co-regularized methods impose pairwise inter-view collaborations. SVM-2K~\cite{SVM-2K} encourages pairwise alignment, while coupling-based MV-LSSVM~\cite{houthuys2018coupling} incorporates pairwise products of the errors into Least-Squares SVM (LS-SVM) leading to complementary views that can offset each other’s errors. Co-regularization methods fail to capture global interactions among multiple views. To address this limitation, $\varrho$TMV-RKM~\cite{houthuys2021tensor} uses tensor-based Restricted Kernel Machines with a joint optimization to create a shared latent space that aligns all views at once. \\
Mumbo~\cite{kocco2011boosting} is a boosting scheme that combines weak learners from each view through sample-weight updates. It promotes a limited form of complementarity by encouraging weaker views to focus on samples that are difficult for the major view. Consequently, its global   collaboration mechanism is mainly centered around the dominant view.

We propose Adaptive Weighted LS-SVM (AW-LSSVM) for multi-view classification, a method that explicitly enforces complementary collaboration across views. Unlike MV-LSSVM and Mumbo, it takes into account the weaknesses of all views when guiding each view’s learning, rather than restricting the interaction to pairwise or dominant-view-centered compensation, and unlike $\varrho$TMV-RKM, which aligns all views in a joint latent space, it enforces global complementarity directly in the original feature spaces through adaptive error-based sample weighting, retaining the distinctiveness of each view. Experiments show that AW-LSSVM achieves strong classification performance on a variety of benchmark multi-view datasets. An added advantage of the proposed method is that raw features remain separated across views, making it naturally suitable for distributed settings such as Vertical Federated Learning (VFL)~\cite{liu2024vertical}, though formal privacy-preserving realization and experiments are left for future work.
\section{Proposed Model}
Multi-view data often exhibit differences in distribution across views, and this diversity allows them to provide complementary information that can be leveraged to enhance model performance. Motivated by this, we propose Adaptive Weighted LS-SVM (AW-LSSVM), an iterative approach for multi-view classification that captures global collaboration and promotes compensation of misclassification errors across views. For each view, this is achieved by adaptively assigning higher weights to samples misclassified by other views. 
 Unlike SVM~\cite{cortes1995support} where only margin-violating samples contribute to the loss, in LS-SVM~\cite{suykens2002least}, each sample has an error variable that measures the deviation from the ideal margin constraint and contributes to the loss. This property is particularly suitable for our adaptive approach. Given a dataset 
$\{(\mathbf{x}_k^{(v)}, y_k)\}_{\scriptscriptstyle k=1,v=1}^{\scriptscriptstyle N,V}$
 with $N$ samples and $V$ views, at each iteration \(t>0\), AW-LSSVM solves the following \textbf{primal} problem for each view \(v\):

\begin{align}\label{eq:AW-LSSM primal}
\min_{\mathbf{w}^{(v,t)},\,b^{(v,t)},\,e^{(v,t)}} \;
& \frac{1}{2}\|\mathbf{w}^{(v,t)}\|^2
 + \frac{\gamma^{(v,t)}}{2}\sum_{k=1}^N (e_{k}^{(v,t)})^2
 + \frac{\rho^{(v,t)}}{2}\sum_{k=1}^N s_{k}^{(v,t-1)} (e_{k}^{(v,t)})^2 \notag \\[3pt]
\text{s.t.}\quad
& y_k\!\left(\mathbf{w}^{(v,t)\mathrm{T}}\varphi^{(v,t)}(\mathbf{x}_k^{(v)}) + b^{(v,t)}\right)
 = 1 - e_{k}^{(v,t)}, \quad \forall k
\end{align}
\noindent where, \(\mathbf{w}^{(v,t)}\) is the weight vector, \(b^{(v,t)}\) is the bias value, \(\varphi^{(v,t)}(\cdot)\) is the feature map, and \(\gamma^{(v,t)} > 0\) is the regularization parameter. While the standard LS-SVM regularization term \(\sum_{k=1}^{N} (e_k^{(v,t)})^2\) penalizes all sample errors uniformly, the proposed global coupling term \(\sum_{k=1}^{N} s_k^{(v,t-1)} (e_k^{(v,t)})^2\), controlled by the hyperparameter \(\rho^{(v,t)}\), links the optimization of each view to the misclassification behavior of all other views through \(s_k^{(v,t-1)}\). This additional term encourages each view-specific model not only to generalize on its own data but also to compensate for the weaknesses of other views by assigning higher weights to samples that were misclassified by them. To induce this error-compensation mechanism, the weights are computed from the aggregated errors of misclassified samples across views. Specifically, the error vectors from the previous iteration are first modified as \(\tilde{\mathbf{e}}^{(v,t-1)}\), retaining only the errors of misclassified samples while setting the rest ($e_k \leq 1$) to zero. The resulting modified errors from other views are then aggregated and incrementally updated to maintain stability across iterations. For this aggregation, we consider two strategies; in addition to simple averaging (Eq. (\ref{eq:ave_aggregation})), we introduce a novel dissimilarity-aware weighted aggregation using Eq.~(\ref{eq:dissim_aggregation}). This strategy leverages the normalized pairwise Euclidean distance between modified error vectors to assign higher importance to views with more dissimilar error patterns.
\begin{equation}\label{eq:ave_aggregation}
s_{k,\text{avg}}^{(v,t-1)}
= \beta^{t-2}\frac{1}{V-1}\sum_{\substack{u=1 \\ u \ne v}}^{V}
(\tilde e_k^{(u,t-1)})^2 + s_k^{(v,t-2)}
\end{equation}
\begin{equation}\label{eq:dissim_aggregation}
  s_{k, dissim}^{(v,t-1)}
  = \beta^{\,t-2}\sum_{u=1}^{V} \frac{\bigl\lVert (\mathbf{\tilde{e}}^{(v,t-1)})^{\circ 2}
    - (\mathbf{\tilde{e}}^{(u,t-1)})^{\circ 2}\bigr\rVert_2}
         {\displaystyle\sum_{u'}\bigl\lVert (\mathbf{\tilde{e}}^{(v,t-1)})^{\circ 2}
    - (\mathbf{\tilde{e}}^{(u',t-1)})^{\circ 2}\bigr\rVert_2} \,
    (\tilde e_{k}^{(u,t-1)})^2 \;+\; s_{k}^{(v,t-2)}
\end{equation}
\noindent with $s_{k}^{(v,0)}=0$ at the initial round \(t=1\), \( ^{\circ}\) denotes element-wise operation, and \(\beta^{(t-2)} \in (0,1)\) is the decay factor that gradually reduces the increments across iterations, serving as an informal stabilizer. Notably, this stabilization is enforced by $\beta$ independently of $\rho$, the kernel, or the cross-view error signals. In practice, the choice of aggregation strategy depends on the structure of inter-view diversity. The average-based variant is preferable when the views are relatively consistent, as it provides a more stable coupling mechanism. The dissimilarity-aware variant is more suitable when the views are less redundant and exhibit complementary information, as it emphasizes informative differences and promotes stronger cross-view compensation. Such inter-view diversity can be assessed using metrics such as Centered Kernel Alignment (CKA)~\cite{cortes2012algorithms}.

By taking the Lagrangian of the primal problem, deriving the KKT optimality conditions, and eliminating the primal variables $\textbf{w}$ and $e$, we obtain the following \textbf{dual} problem:
\begin{equation}\label{eq:aw_lssvm_dual}
\left[
\begin{array}{c|c}
0 & \mathbf{y}^{(v,t)^{T}} \\ \hline
\mathbf{y}^{(v,t)} & \mathbf{\Omega}^{(v,t)} +  \Lambda^{(v,t)}
\end{array}
\right]
\,
\left[
\begin{array}{c}
b^{(v,t)} \\ \hline
\boldsymbol{\alpha}^{(v,t)}
\end{array}
\right]
=
\left[
\begin{array}{c}
0\\ \hline
\mathbf{1}_{N}
\end{array}
\right]
\end{equation}

%the operator \( \circ \) denotes element-wise multiplication.
\noindent where, \(\mathbf{\Lambda}^{(v,t)}=\operatorname{diag}\!\bigl((\gamma^{(v,t)}+\rho^{(v,t)}\,s_{1}^{(v,t-1)})^{-1},\dots,(\gamma^{(v,t)} +\rho^{(v,t)}\,s_{N}^{(v,t-1)})^{-1}\bigr)\in \mathbb{R}^{N\times N} \) is a sample-specific inverse weighting matrix, $\mathbf{\Omega}^{(v,t)}$ is a labeled kernel matrix with elements $\Omega_{ij} = y_i y_j K(\mathbf{x}_i^{(v)}, \mathbf{x}_j^{(v)})$, using a kernel function \(K:\mathbb{R}^{d_v}\times\mathbb{R}^{d_v}\rightarrow\mathbb{R}\), where \(d_v\) denotes the dimensionality of view \(v\). Solving Eq. (\ref{eq:aw_lssvm_dual}) yields dual variables $\boldsymbol{\alpha}^{(v,t)}\in \mathbb{R}^{N}$ and a bias variable $b^{(v,t)} \in \mathbb{R}$ to compute the error vector of training samples for the next iteration as follows: 
\begin{equation}\label{eq:error}
\mathbf{e}^{(v,t)} = \mathbf{y} - (\mathbf{\Omega}^{(v,t)}\boldsymbol{\alpha}^{(v,t)} + b^{(v,t)}\mathbf{1})
\end{equation}
For multiclass datasets, one-vs-all label encoding produces multiple binary AW-LSSVM subproblems for each view. After the optimal number of iterations \(T\), the prediction for a new test sample $\mathbf{x}^*=\{\mathbf{x}^{*(v)}\}_{v=1}^V$ is obtained using Eq.~(\ref{eq:final_score}), which aggregates the soft decision values of all view-specific classifiers, each computed via the decision function in Eq.~(\ref{eq:decision_func}):
\begin{equation}\label{eq:decision_func}
f^{(v,T)}(\mathbf{x}^{*(v)})=
\sum_{k=1}^{N}\alpha_k^{(v,T)} y_k
K\!\left(\mathbf{x}^{*(v)},\mathbf{x}_k^{(v)}\right)+b^{(v,T)}
\end{equation}

\begin{equation}\label{eq:final_score}
\hat{y}(\mathbf{x}^*)=
\operatorname{sign}\!\left(
\frac{1}{V}\sum_{v=1}^{V} f^{(v,T)}\!\left(\mathbf{x}^{*(v)}\right)
\right)
\end{equation}

  \begin{figure}
\includegraphics[width=\textwidth, trim=1 2cm 0 2cm, clip]{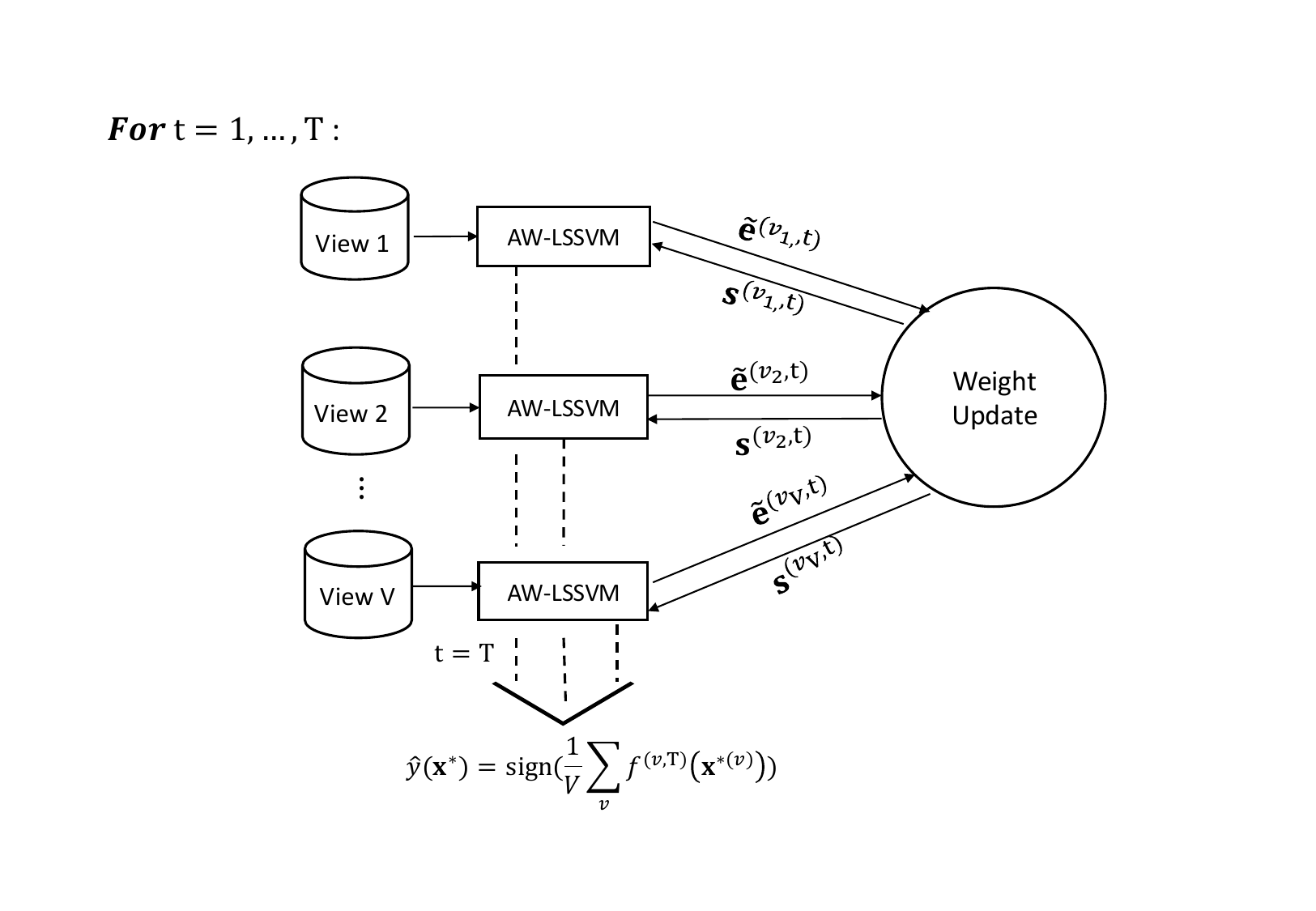}
\caption{Illustration of the proposed algorithm. At each iteration, view-specific misclassified error vectors $\tilde{\mathbf{e}}^{(v,t)}$ are aggregated to update sample weights for the subsequent iterations.}
\label{AW-LSSVM}
\end{figure}

Fig.~\ref{AW-LSSVM} illustrates the overall framework of the proposed method. As demonstrated, this algorithm's ability to keep raw features isolated across views suggests potential applicability in VFL, though privacy guarantees remain outside the scope of this paper. The detailed training and inference procedures are summarized in Algorithm~\ref{alg:awlssvm}.
\newcommand{\algsubsection}[1]{%
  \par\smallskip
  \noindent\hspace*{2.2em}\colorbox{gray!20}{\strut\textbf{#1}}\par
}

\begin{algorithm*}[t]
\caption{Multi-View Classification by AW-LSSVM}
\label{alg:awlssvm}

\algsubsection{Training}
\setcounter{AlgoLine}{0}
\KwIn{$\mathcal{D}=\{(\mathbf{x}_k^{(v)}, y_k)\}_{k=1,v=1}^{N,V}$, $K(\cdot,\cdot)$, $\gamma,\rho,\beta$, $T$, $\mathcal{A}\in\{\mathrm{avg},\mathrm{dissim}\}$}

Initialize $s_k^{(v,0)}=0$\;

\For{$t=1$ \KwTo $T$}{
    \ForEach{$v=1,\dots,V$}{
        Solve AW-LSSVM using $s_k^{(v,t-1)}$\tcp*[r]{Eq.~\eqref{eq:aw_lssvm_dual}}
        Compute error vector $\mathbf{e}^{(v,t)}$\tcp*[r]{Eq.~\eqref{eq:error}}
        $\tilde{\mathbf{e}}^{(v,t)} \gets \mathbf{e}^{(v,t)} \text{ with entries } \tilde{e}_k^{(v,t)} = 0 \text{ whenever } e_k^{(v,t)} \le 1$ \;
    }
    \ForEach{$v=1,\dots,V$}{
        \If{$\mathcal{A}=\mathrm{avg}$}{
            Update $s_{k,\mathrm{avg}}^{(v,t)}$\tcp*[r]{Eq.~\eqref{eq:ave_aggregation}}
        }
        \If{$\mathcal{A}=\mathrm{dissim}$}{
            Update $s_{k,\mathrm{dissim}}^{(v,t)}$\tcp*[r]{Eq.~\eqref{eq:dissim_aggregation}}
        }
    }
}
\KwOut{trained model $\{(\boldsymbol{\alpha}^{(v,T)}, b^{(v,T)})\}_{v=1}^{V}$}
\setcounter{AlgoLine}{1}
\algsubsection{Inference}
\setcounter{AlgoLine}{0}
\KwIn{Test sample $\mathbf{x}^*=\{\mathbf{x}^{*(v)}\}_{v=1}^V$, trained model $\{(\boldsymbol{\alpha}^{(v,T)}, b^{(v,T)})\}_{v=1}^{V}$}

\ForEach{$v=1,\dots,V$}{
    Compute $f^{(v,T)}(\mathbf{x}^{*(v)})$\tcp*[r]{Eq.~\eqref{eq:decision_func}}
}
Compute $\hat{y}(\mathbf{x}^*)$\tcp*[r]{Eq.~\eqref{eq:final_score}}
\KwOut{$\hat{y}(\mathbf{x}^*)$}
\end{algorithm*}
\begin{table*}[t]
\centering
\footnotesize
\setlength{\tabcolsep}{3pt}
\renewcommand{\arraystretch}{1.1}
\caption{Properties of the multi-view datasets used in experiments.}
\label{tab:dataset-stats}
\begin{minipage}{0.475\textwidth}
\centering
\begin{tabular}{lccc}
\hline
Dataset & Samples & Views & Classes \\
\hline
3Sources     & 169  & 3  & 6  \\
YT           & 500  & 3  & 7  \\
NUSWIDE      & 2400 & 5  & 12 \\
Prokaryotic  & 551  & 3  & 4  \\
ProteinFold  & 694  & 12 & 27 \\
\hline
\end{tabular}
\end{minipage}
\hfill
\begin{minipage}{0.475\textwidth}
\centering
\begin{tabular}{lccc}
\hline
Dataset & Samples & Views & Classes \\
\hline
Caltech       & 1474 & 6 & 7  \\
Flower        & 1360 & 7 & 17 \\
ACM           & 3025 & 5 & 3  \\
Cora          & 2708 & 4 & 7  \\
OutdoorScene  & 2688 & 4 & 8  \\
\hline
\end{tabular}
\end{minipage}
\end{table*}
\section{Experiments}
In this section, we apply the proposed AW-LSSVM model to several benchmark multi-view datasets and compare its performance with other kernel-based multi-view classification methods. The code will be made publicly available upon publication.

\subsection{Datasets and Baseline Models}
We conducted experiments on ten benchmark multi-view datasets\footnote{Data are available at \url{https://github.com/ChuanbinZhang/Multi-view-datasets}}~\cite{3sources,YT,Chua2009NUSWIDE,prokaryotic,proteinFold,Flowers,ACM,sen2008Cora,outdoorscene}, whose statistical properties are summarized in Table~\ref{tab:dataset-stats}. Due to the large size of the YT (YouTube video games) dataset~\cite{YT}, stratified downsampling was applied to this dataset to subsample 500 instances from seven most frequent categories. From its multiple high-dimensional views, two auditory views (Spectrogram and MFCC) and one visual view (HOG) were chosen.

 We compare AW-LSSVM with the baseline models \textbf{BSV}, \textbf{Early Fusion}, and \textbf{Late Fusion} as well as the  state-of-the-art kernel-based classifiers \textbf{MV-LSSVM}~\cite{houthuys2018coupling}, \textbf{EasyMKL}~\cite{aiolli2015easymkl}, \textbf{Mumbo}~\cite{kocco2011boosting}, and \textbf{$\varrho$TMV-RKM}~\cite{houthuys2021tensor}. \textbf{BSV} applies LS-SVM on each view separately, reporting the best-performing view. \textbf{Early Fusion} concatenates all view features before training a single LSSVM, while \textbf{Late Fusion} combines predictions of view-specific LS-SVM classifiers via majority voting. 
 
\subsection{Model Selection}
For robust hyperparameter tuning, each dataset is divided into three distinct train–test splits, and for each training split, hyperparameter optimization is carried out using dual annealing combined with three-fold cross-validation.

 In order to reduce computational complexity, for all one-vs-all binary subproblems of a given view, the same hyperparameters are shared. In AW-LSSVM, the same hyperparameter values are also used across all iterations. Although separate kernel widths and regularization parameters could in principle be tuned for each view, we use a shared RBF kernel width \(\sigma\) and a common regularization parameter \(\gamma\) across views for all methods except BSV. The remaining method-specific hyperparameters of each baseline, as well as those of AW-LSSVM (\(\rho^{(v)}\), \(T\) with a maximum of five, and \(\beta\)), are tuned accordingly. Algorithm~\ref{alg:model_selection} describes the complete model selection procedure used in the experiments.

 \begin{algorithm*}[t]
\caption{Model Selection}
\label{alg:model_selection}

\KwIn{$\mathcal{D}=\{(\mathbf{x}_k^{(v)}, y_k)\}_{k=1,v=1}^{N,V}$, $K(\cdot,\cdot)$, search spaces for $\gamma$, $\sigma$, $\rho^{(v)}$, $\beta$, $T$}
Split $\mathcal{D} \rightarrow \{\mathcal{D}_i^{\text{train}}, \mathcal{D}_i^{\text{test}}\}_{i=1}^{3}$\;

\For{$i=1,2,3$}{
    $(\gamma_i^*, \sigma_i^*, \rho_i^{(v)*}, \beta_i^*, T_i^*)
    \leftarrow \text{DA \& CV$_{3}$, Alg.~\ref{alg:awlssvm} on } \mathcal{D}_i^{\text{train}}$\;

    $\hat{\mathbf y}^{test}_i \leftarrow \text{Alg.~\ref{alg:awlssvm}}\bigl( \mathcal{D}_i^{\text{test}}, K, \gamma_i^*, \sigma_i^*, \rho_i^{(v)*}, \beta_i^*, T_i^*\bigr)$\;
}

\KwOut{$\hat{\mathbf y}^{test}_1,\hat{\mathbf y}^{test}_2,\hat{\mathbf y}^{test}_3$}

\end{algorithm*}

\subsection{Results}
\begin{table}[ht]
\centering
\large
\setlength{\tabcolsep}{4pt}
\renewcommand{\arraystretch}{1.15}
\caption{Mean and std. of balanced accuracy on three hold-out test splits. \textbf{Bold} values indicate the highest score. In the Flower dataset, since the kernels have already been precomputed, the Mumbo  model is not applicable.}
\label{tab:results}
\begin{adjustbox}{max width=\textwidth}
\begin{tabular}{l c c c c c c }
\toprule
& \multicolumn{5}{c}{\textbf{Datasets}} \\
\midrule
\textbf{Method} & 3Sources & YT & NUSWIDE & Prokaryotic & ProteinFold \\
\midrule
BSV & 72.50{\small (±12.31)} & 79.66{\small (±2.18)} & 38.89{\small (±2.41)} & 66.88{\small (±5.90)} & 59.76{\small (±0.53)} \\
Early Fusion & 67.13{\small (±4.24)} & 73.17{\small (±0.16)} & 49.03{\small (±3.46)} & 66.56{\small (±6.85)} & 58.22{\small (±1.93)} \\
Late Fusion & 63.89{\small (±2.41)} & 68.83{\small (±2.15)} & 43.54{\small (±1.16)} &66.31 {\small (±8.73)} & 47.96{\small (±2.99)} \\
MV-LSSVM & 68.98{\small (±9.85)} & 77.14{\small (±0.14)} & 47.64{\small (±2.04)} & 62.17{\small (±2.25)} & 57.37{\small (±3.45)} \\
EasyMKL & 81.48{\small (±9.25)} & 77.82{\small (±1.52)} & \textbf{49.58}{\small (±3.15)} & 67.08{\small (±2.66)} &\textbf{68.09} {\small (±3.72)} \\
Mumbo & 55.56{\small (±8.67)} & 76.21{\small (±4.40)} & 39.31{\small (±1.39)} & 63.39{\small (±4.39)} & 46.44{\small (±1.20)} \\
$\varrho$TMV-RKM & 89.04{\small (±2.10)} & 78.73{\small (±4.03)} & 48.26{\small (±1.70)} & 79.58{\small (±2.11)} & 67.74{\small (±5.19)}\\

\rowcolor{gray!20} AW-LSSVM$_{avg}$ & 89.73{\small (±1.68)} & 78.62{\small (±2.10)}& 47.01{\small (±1.58)} &83.31 {\small (±0.71)} & 65.79{\small (±1.49)}  \\
\rowcolor{gray!20} AW-LSSVM$_{dissim}$ & \textbf{91.07}{\small (±0.71)} & \textbf{80.57}{\small (±2.22)} & 48.33{\small (±1.78)} & \textbf{83.77}{\small (±1.38)} & 64.99{\small (±1.40)} \\
\midrule
& Caltech & Flower & ACM & Cora & OutdoorScene \\
\midrule
BSV & 93.14{\small (±3.20)} & 75.74{\small (±2.41)} & 91.91{\small (±0.56)} & 76.22{\small (±0.76)} &  88.04 {\small (±0.38)}\\
Early Fusion & 94.37{\small (±1.68)} & 88.36{\small (±1.29)} & 93.04{\small (±0.77)} & 80.95{\small (±1.83)} & 89.69{\small (±0.61)}\\
Late Fusion & 88.81{\small (±4.90)} & 83.95{\small (±1.29)} & 89.78{\small (±1.00)} & 39.19{\small (±2.37)} & 89.55{\small (±0.22)}\\
MV-LSSVM & 91.91{\small (±0.75)} & 89.22{\small (±1.49)} & 92.72{\small (±1.59)} & 77.46{\small (±3.04)} &89.91 {\small (±1.09)}\\
EasyMKL & 92.96{\small (±2.06)} & 88.60{\small (±1.10)} & 93.60{\small (±1.44)} & 79.76{\small (±5.51)} & 89.58{\small (±0.58)}\\
Mumbo & 93.03{\small (±0.43)} & -- & 89.37{\small (±1.95)} & 71.52{\small (±2.72)} & 85.52 {\small (±0.45)}\\
$\varrho$TMV-RKM & 95.12{\small (±2.70)} & 88.48{\small (±1.29)} & 93.22{\small (±1.21)} & \textbf{85.02}{\small (±0.14)} & 90.24{\small (±0.99)}\\

\rowcolor{gray!20} AW-LSSVM$_{avg}$ & \textbf{96.23}{\small (±3.84)} &\textbf{90.20} {\small (±1.29)} &94.90{\small (±0.74)} &84.26 {\small (±2.77)} & \textbf{90.31}{\small (±0.42)}\\
\rowcolor{gray!20} AW-LSSVM$_{dissim}$ & 95.43{\small (±4.24)} & 89.95{\small (±1.12)} & \textbf{95.04}{\small (±0.80)} & 82.80{\small (±1.73)} & 90.02{\small (±0.99)}\\
\bottomrule
\end{tabular}
\end{adjustbox}
\end{table}

\begin{figure}
\includegraphics[width=\textwidth]{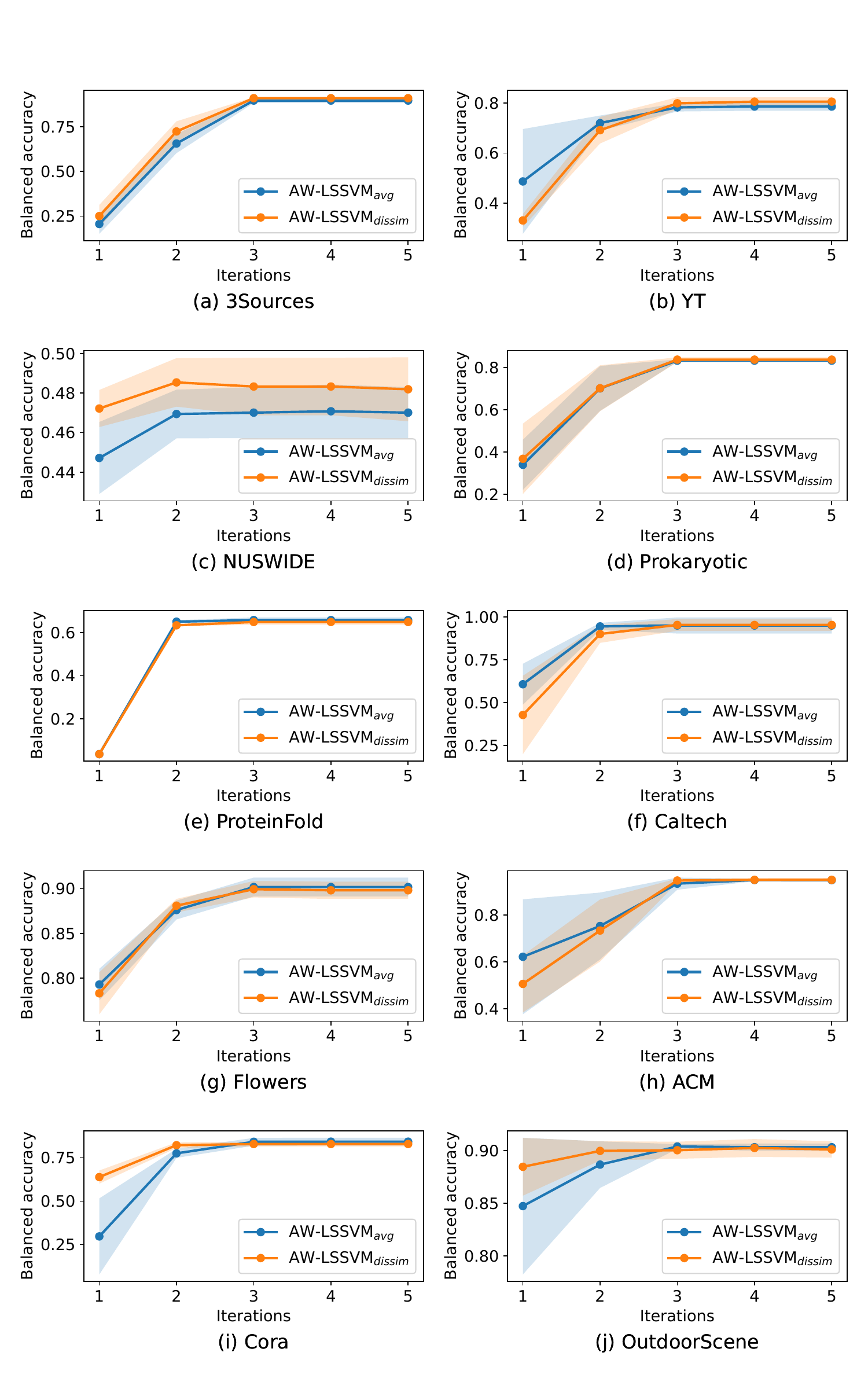}
\caption{Mean and standard deviation of balanced accuracy for AW-LSSVM$_{\text{avg}}$ and AW-LSSVM$_{\text{dissim}}$ over three test splits with respect to iterations $t$ across different datasets. For clarity, note that the y-axis scales are not consistent across subplots.} 
\label{convergence_plots}
\end{figure}

The final results in Table~\ref{tab:results} report the balanced accuracies on three hold-out test splits using the corresponding optimal hyperparameters for each benchmark multi-view dataset. The last two rows correspond to the proposed AW-LSSVM model with two aggregation strategies: AW-LSSVM$_{avg}$ and AW-LSSVM$_{dissim}$, which use average-based and dissimilarity-aware error aggregation, respectively, as defined in Eq.~(\ref{eq:ave_aggregation}) and Eq.~(\ref{eq:dissim_aggregation}).
The results show that AW-LSSVM performs strongly across the benchmark datasets and achieves the highest mean balanced accuracy on seven out of ten datasets, highlighting the benefit of effective sample-weighted global error compensation among views. Fig.~\ref{convergence_plots} shows the mean and standard deviation of the test balanced accuracy of AW-LSSVM$_{\text{avg}}$ and AW-LSSVM$_{\text{dissim}}$ over five iterations across three different test splits\footnote{Note that, although the results in Table~\ref{tab:results} are reported using the tuned iterations \(T\), the models in Fig.~\ref{convergence_plots} were trained for five iterations on all datasets and splits to get a consistent visualization.}. As shown, three iterations (with \(t=1\) corresponding to the initial LS-SVM round) generally yield the highest test performance. This suggests that the method has limited sensitivity to the choice of \(T\), and that for datasets with similar sizes and properties, a small fixed value such as \(T=3\) may already be sufficient in practice without requiring extensive tuning. This early convergence can improve communication efficiency in FL settings, where secure data exchange among parties holding different views is essential but costly.

Examining the individual performances of the LS-SVM view-specific classifiers in Fig.~3 for ProteinFold and Cora, where AW-LSSVM does not achieve the best overall result, shows that the majority of views in ProteinFold and half of the views in Cora are weak, with balanced accuracy below 50\%. These weak views are either insufficiently informative or noisy. Although AW-LSSVM benefits from cross-view interaction and therefore outperforms most baselines on these datasets, its error-coupling mechanism may become less effective when several views are weak. In such cases, strong views may be guided by unreliable error signals from weak views, while weak views may be encouraged to focus on hard samples that they cannot model well. This can degrade some view-specific classifiers and limit the overall gain under uniform score aggregation. 

\begin{figure}
\includegraphics[width=\textwidth]{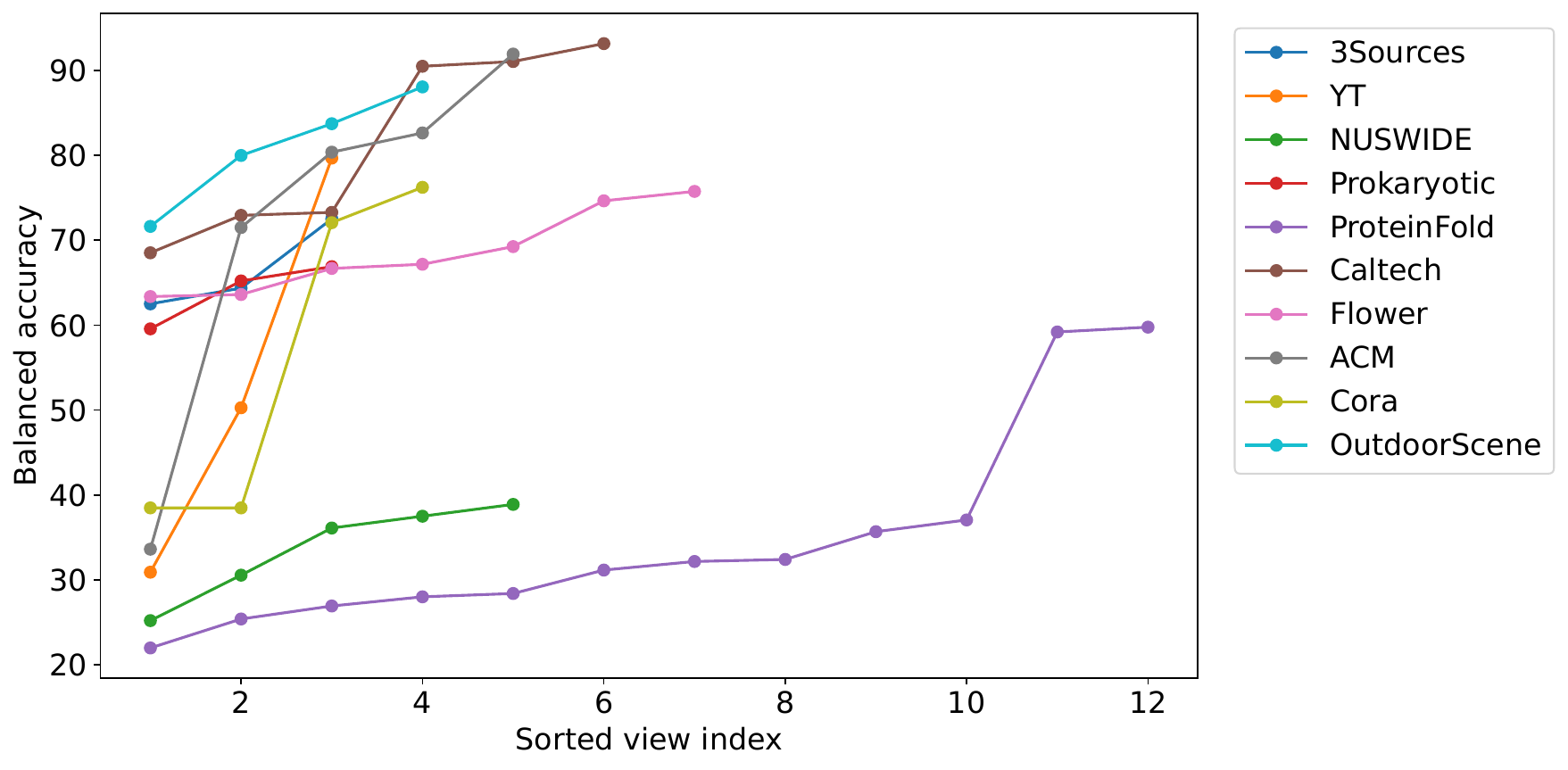}
\caption{Sorted single-view LS-SVM balanced accuracy scores for each dataset. Each curve shows the ordered performance of the individual views, highlighting the distribution of weak and strong views within each dataset.} 
\label{single views}
\end{figure}

This also helps explain the relative behavior of the two AW-LSSVM variants. Overall, AW-LSSVM\(_{dissim}\) performs only slightly better than AW-LSSVM\(_{avg}\),
with mean balanced accuracies of 82.20 and 82.04 over all datasets, respectively,
corresponding to a gain of 0.16. However,  this advantage becomes more noticeable when excluding ProteinFold and Cora, where the gain increases to 0.48. This is consistent with the discussion above, since the dissimilarity-aware variant places greater emphasis on views with more distinct error patterns, which in datasets containing both strong and weak views may undesirably amplify the influence of weak or noisy ones. In general, the dissimilarity-aware variant can therefore be used as the default strategy, except for datasets with a combination of strong and weak views in which a large proportion of the views are weak.

\begin{figure}
\includegraphics[width=\textwidth]{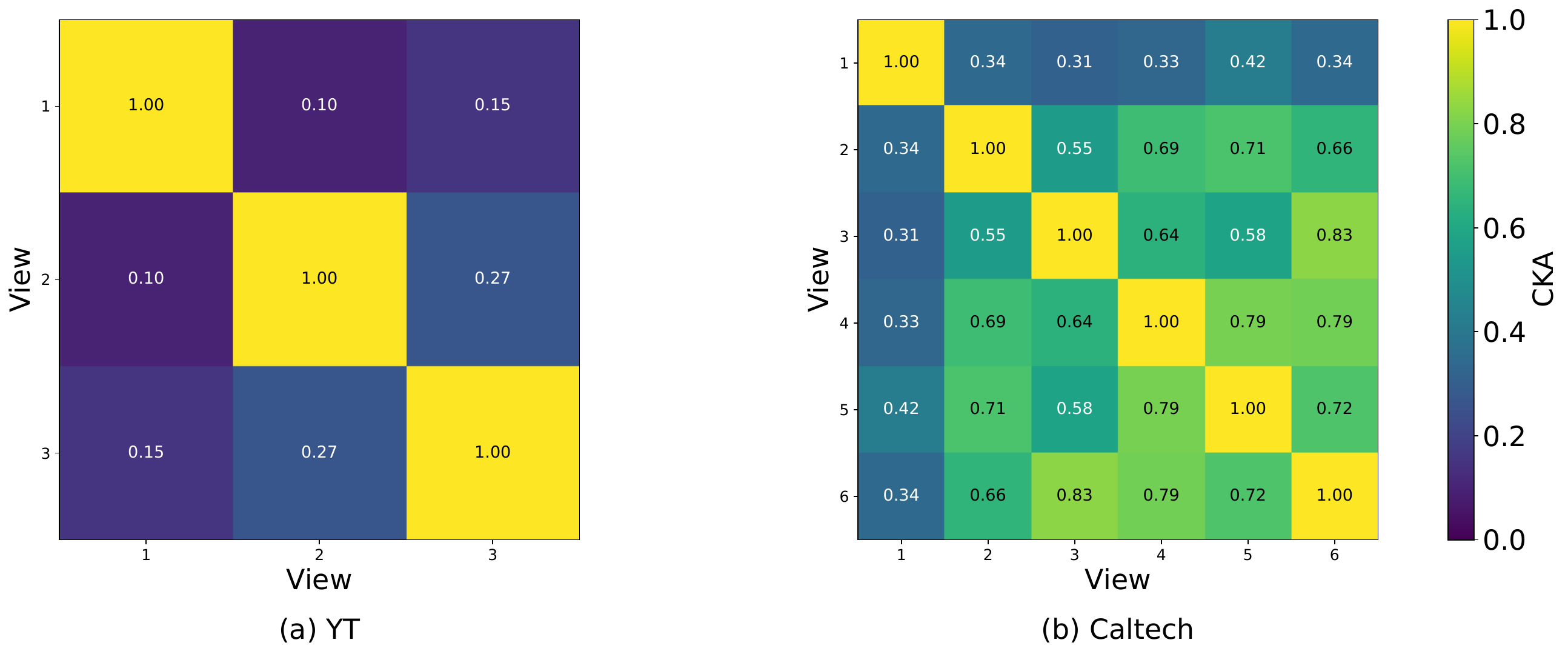}
\caption{Pairwise centered kernel alignment (CKA) between views for the Caltech and YT datasets, computed using RBF kernels with the default kernel width.}
\label{cka}
\end{figure}

CKA can further support the choice between the two variants as an indicator of inter-view similarity. Fig.~\ref{cka} shows pairwise CKA heatmaps for the Caltech and YT datasets, which exhibit the largest gains of AW-LSSVM\(_{avg}\) over AW-LSSVM\(_{dissim}\) and of AW-LSSVM\(_{dissim}\) over AW-LSSVM\(_{avg}\), respectively. The CKA values are computed using RBF kernels with the default kernel width. The relatively high average pairwise CKA in Caltech (0.58) indicates that the views are more similar and aligned, which is consistent with the stronger performance of the average-based variant. The lower average pairwise CKA in YT (0.17) indicates a more diverse view structure, which is consistent with the stronger performance of the dissimilarity-aware variant.

EasyMKL, which performs weighted kernel averaging across views, and \(\varrho\)TMV-RKM, which projects the views into a shared subspace, may be more effective on datasets with a large proportion of weak views (e.g., NUSWIDE, ProteinFold, and Cora), as they can reduce the impact of view-specific noise or overall lack of informativeness. Nevertheless, replacing the uniform aggregation in the decision rule of Eq.~(7) with a weighted averaging scheme based on measures such as the error covariance matrix may further improve AW-LSSVM. In addition, kernel-level aggregation is compatible with our framework, allowing earlier fusion strategies to be incorporated when beneficial, for example on NUSWIDE, where the strong Early Fusion result suggests that the views can interact effectively at an earlier stage.

\subsection{Statistical analysis}

For statistical analysis of the performance improvement of AW-LSSVM over baseline methods, we conducted pairwise Wilcoxon signed-rank~\cite{demvsar2006statistical} tests across all datasets. In particular, when compared against BSV, Early Fusion, Late Fusion, MV-LSSVM, EasyMKL, and Mumbo, the Wilcoxon statistics indicate that our method significantly outperforms these methods, with $p$-values below $0.05$ in all cases. The comparison with $\varrho$TMV-RKM produces $T = 11.0$ and $p$-value $= 0.1$, indicating no significant difference between the two methods. Overall, these results show that AW-LSSVM delivers statistically significant gains over the considered baseline methods, while performing comparably to the strongest competing method, $\varrho$TMV-RKM.

\section{Conclusion and Future Work}
This paper introduced AW-LSSVM, a novel kernel-based method for multi-view classification that explicitly promotes complementary learning across all views through a global coupling term. This term adaptively assigns additional sample weights to each view, guiding it to compensate for samples misclassified by the other views. The weights are computed from the aggregated misclassification errors in previous iterations. For this aggregation, besides a uniform view-error aggregation strategy, we also proposed a dissimilarity-aware variant for more coordinated error compensation, in which each view is encouraged to place greater emphasis on errors from views with which it is less redundant. Experimental results on ten benchmark datasets show that AW-LSSVM improves performance over state-of-the-art multi-view methods that either do not explicitly enforce any particular form of collaboration across views or model collaboration only in a pairwise manner.
By relying solely on view-level error and decision information, the proposed method keeps raw features separated across views and is therefore architecturally suitable for distributed multi-view settings such as VFL. Nevertheless, formal privacy guarantees are not established in this work. Future work will investigate federated implementations of AW-LSSVM and evaluate their communication, security, and privacy properties. Extending AW-LSSVM to a semi-supervised setting is another promising research direction.
\begin{credits}
\subsubsection{\ackname} We would like to thank the Flemish Government under the Onderzoeksprogramma Artificiele Intelligentie (AI) Vlaanderen programme for funding this research.
\subsubsection{\discintname} The authors have no competing interests to declare.
\end{credits}

% ---- Bibliography ----

\bibliographystyle{splncs04}
\bibliography{bib}

\end{document}